\documentclass{article}

 \usepackage[preprint]{neurips_2025}

\usepackage[utf8]{inputenc} 
\usepackage[T1]{fontenc}    
\usepackage{hyperref}       
\usepackage{url}            
\usepackage{booktabs}       
\usepackage{amsfonts}       
\usepackage{nicefrac}       
\usepackage{microtype}      
\usepackage{xcolor}         
\usepackage{amsmath}
\usepackage{graphicx}
\usepackage{diagbox}
\usepackage{subfigure}
\usepackage{makecell}
\usepackage{tabularx}

\title{An Advanced Convolutional Neural Network for Bearing Fault Diagnosis under Limited Data}

\author{%
  Shengke Sun \\
  Nanjing University of Science and Technology\\
  \texttt{sunshengke@njust.edu.cn} \\
  \And
  Shuzhen Han \\
  Tiangong University \\
  \texttt{hanshuzhen@tiangong.edu.cn} \\
  \AND
  Ziqian Luan \\
  Xidian University \\
  \texttt{tyuwwe@foxmail.com} \\
  \And
  Xinghao Qin \\
  Tianjin University \\
  \texttt{2111640317@tiangong.edu.cn} \\
  \And
  Jiao Yin \\
  Victoria University \\
  \texttt{jiao.yin@vu.edu.au} \\
  \And
  Zhanshan Zhao \\
  Tiangong University \\
  \texttt{zhzhsh127@163.com} \\
  \And
  Jinli Cao \\
  La Trobe University \\
  \texttt{j.cao@latrobe.edu.au} \\
  \And
  Hua Wang \\
  Victoria University \\
  \texttt{hua.wang@vu.edu.au} \\
}

\begin{document}

\maketitle

\begin{abstract}
  In the area of bearing fault diagnosis, deep learning (DL) methods have been widely used recently. However, due to the high cost or privacy concerns, high-quality labeled data are scarce in real world scenarios. While few-shot learning has shown promise in addressing data scarcity, existing methods still face significant limitations in this domain. Traditional data augmentation techniques often suffer from mode collapse and generate low-quality samples that fail to capture the diversity of bearing fault patterns. Moreover, conventional convolutional neural networks (CNNs) with local receptive fields makes them inadequate for extracting global features from complex vibration signals. Additionally, existing methods fail to model the intricate relationships between limited training samples. To solve these problems, we propose an advanced data augmentation and contrastive fourier convolution framework (DAC-FCF) for bearing fault diagnosis under limited data. Firstly, a novel conditional consistent latent representation and reconstruction generative adversarial network (CCLR-GAN) is proposed to generate more diverse data. Secondly, a contrastive learning based joint optimization mechanism is utilized to better model the relations between the available training data. Finally, we propose a 1D fourier convolution neural network (1D-FCNN) to achieve a global-aware of the input data. Experiments demonstrate that DAC-FCF achieves significant improvements, outperforming baselines by up to 32\% on case western reserve university (CWRU) dataset and 10\% on a self-collected test bench. Extensive ablation experiments prove the effectiveness of the proposed components. Thus, the proposed DAC-FCF offers a promising solution for bearing fault diagnosis under limited data.
\end{abstract}

\section{Introduction}
\label{sec:introduction}

As a pivot component of industrial machinery, the health condition of rolling bearings is crucial for the normal operation of industrial machinery~\cite{1}. When rolling bearings fail, it will inevitably affect the stability and continuity of industrial production~\cite{2,3}. Thus, there is a strong demand to develop a reliable and effective rolling bearing fault detection model.

To solve the problem of bearing fault diagnosis, extensive methods based on signal processing such as singular value decomposition\cite{4}, variational mode decomposition\cite{5} and wavelet transform\cite{6} have been proposed. However, these methods inevitably need a careful handcraft feature design, preventing the models from moving towards greater automation and end-to-end training. Recently, due to its flexible feature extraction process, many researchers turned to develop the bearing fault diagnosis models using Deep Learning (DL) methods. Rafiee et al.\cite{7} first utilized a simple MLP for the detection of bearing faults, achieving a high accuracy. To reduce the high computational complexity of fully connected layers, Janssens et al.\cite{8} proposed a convolutional layer for dimensional reduction and feature extraction. To attain a better representation space of bearing signals, Zhang et al.\cite{9} proposed a 5-layer convolutional networks that improved the accuracy to over 90\%. Another issue lies on the path is the overlapping and noisy nature of acoustic signals when sampling, the noise will affect the feature extraction process of neural networks, leading to a bad performance. To solve this problem, Zhang et al.\cite{10} proposed a wide convolution kernel to enhance the receptive field of CNNs for better feature representation. Chen et al.\cite{11} proposed a multi-scale convolution layer for the multi-granularity feature extraction of different parts of signals. Han et al.\cite{han2024deep} combined multi-scale feature extraction with attention mechanism to adaptively fuse features with less noise. Zhong et al.\cite{zhong2024residual} further proposed a residual denoising and multi-scale weighted domain adaptation to enhance the performance in noise domain adaptation tasks.

However, it is well known that DL methods are data-hungry, which need a massive number of training samples to get a high accuracy. While in real world scenarios, due to data privacy or high cost of collecting data, only a small portion of data is available. DL models trained on these limited data usually suffer from overfitting, leading to a bad generalization performance. To overcome this issue, Hu et al.\cite{12} proposed a data augmentation method using Generative Adversarial Networks (GANs) to generate more samples for training, other methods such as modifying network architecture \cite{13,14,add1,add2,add3}, using transfer learning methods\cite{16,wh1} are utilized to relieve the overfitting problem. Although these methods attains some degrees of improvement, existing methods still have significant problems. It is well known that training a GAN is unstable, it requires a careful design of hyperparameters and architecture. Training a GAN often suffers from mode collapse, which the model only generates single type of data. Besides, a critical assumption of transfer learning paradigm is that the label distributions of source domain and target domain have to be the same, while in practice it is not always the case. The frequency domain filter FCNN\cite{20} is only applicable to 2D images, its potential on 1D signals has not been fully explored yet.

Thus there is a significant gap between the ideal and actual conditions. The augmented data generated by conventional GANs usually provide less informative features, and models trained with conventional feature extraction methods tend to extract single type of fault feature, and ignore the intra-relationship within the training data, thus decreasing the reliability of the model. Recently, a new paradigm named CLR-GAN\cite{CLR_GAN} is proposed to make GAN more stable, however, CLR-GAN is an unconditional generative models that can not generate data points according to the give label. Also, the limited receptive field of CNNs preventing the model's global awareness of data, which is a critical property for fault diagnosis under limited data. These issues limit the practical application of the above fault diagnosis models.

Therefore, it is important to develop a model that dives into the deeper correlations of the available data and generates the data more efficiently. In this paper, a new framework named data augmentation and contrastive fourier convolutional framework (DAC-FCF) is proposed, we design a conditional consistent latent representation generative adversarial networks (CCLR-GAN) to generate data points with high fidelity, we also design a 1D Fourier Convolutional Neural Network (1D-FCNN) to extract from both the time domain signal and the frequency domain signal, enhancing the model's global awareness of the available training data. The contribution of this proposed model is listed as follows: 
\begin{enumerate} 
\item We propose a novel framework, Data Augmentation and Contrastive Fourier Convolutional Framework (DAC-FCF), to tackle the challenge of bearing fault diagnosis under limited data conditions. Specifically, DAC-FCF employs a contrastive learning-based joint optimization mechanism to effectively capture both inter- and intra-relationships within the training data. Experimental results show that DAC-FCF improves performance by over 10\% compared to recently proposed methods, demonstrating its effectiveness in data-limited scenarios.
\item We propose a Conditional CLR-GAN (CCLR-GAN) with a specially designed cascade cross-attention module to effectively incorporate label information into augmented features and generate samples more stably than conventional methods like CLR-GAN. Experimental results show that the generated data significantly enhances the diversity of the training dataset, leading to the greatest performance improvement among tested approaches.
\item We design a 1D-FCCN by introduing an adaptive convolutional stride in the global-aware path to the FCNN, tailored for one-dimensional vibration signals of bearing fault diagnosis. We enable FCNN to extract features on 1D signals, greatly improving the model's global feature extraction capability. 
\end{enumerate}

The following sections are arranged as follows: In Sec.\ref{formulation}, we formulate the problem of bearing fault diagnosis under limited data and give a brief introduction to generative adversarial networks, contrastive learing and fourier transform. In Sec.\ref{sec:method}, we give a detailed explanation of the components for the proposed DAC-FCF. In Sec.\ref{sec:exp}, extensive experiments are carried out to validate the effectiveness of the proposed method.
Finally, we summarize the contribution and future works of the proposed method is Sec.\ref{sec:conclusion}.

\section{Problem Formulation and Preliminary Theory}
\label{formulation}
In this section, we give a detailed introduction to the bearing fault diagnosis problem under limited data and provide the preliminary knowledge of the proposed DAC-FCF. 
\subsection{Bearing Fault Diagnosis under Limited Data}
\label{sec:problem_formulation}
In this paper, we focus on mitigating the problem of low fault detection accuracy of bearings under limited data. Given a training dataset $D^{train}:=\{(x^{train}_i,y^{train}_i)|i=1,2,\dots , n^{train}\}$ and a testing dataset $D^{test}:=\{(x^{test}_j)|j=1,2,\dots , n^{test}\}$, where $x^{train}$,$x^{test}$ are the bearing vibration signals and $y^{train}\in \mathbb{R}^{R}$ includes $R$ health states of a bearing.  

The objective of bearing fault diagnosis under limited data is to predict $\hat{y}^{test}$ when $ n^{train} \ll 
 n^{test}$. The core idea of neural networks is to find a representation space where the boundary between normal bearings and fault bearings is clear. However, due to the shortage of data, the conventional model tends to memorize each training sample instead of finding a proper representation space.

 Under such conditions, conventional DL models struggle to learn discriminative feature representations. For example, CNNs with local receptive fields may fail to capture long-range dependencies in vibration signals, while data augmentation techniques like random cropping or noise injection provide limited diversity gains. Furthermore, transfer learning approaches often degrade in performance when the source and target domains exhibit distribution shifts (e.g., different machinery operating conditions).

\subsection{Generative Adversarial Networks}
\label{sec:gan_intro}
Data augmentation methods have been proved a solid solution to data shortage during training. Compared with conventional pixel space augmentation methods (random crop, horizontal flip, Gaussian blur,$\ldots$), generating new data samples using generative models draws more and more attention due to its ability to generative more diverse images and make the data distribution more complete, thus making the model generalize better on the test dataset.

\begin{figure}[!h]
  \centering
  \includegraphics[width=\textwidth]{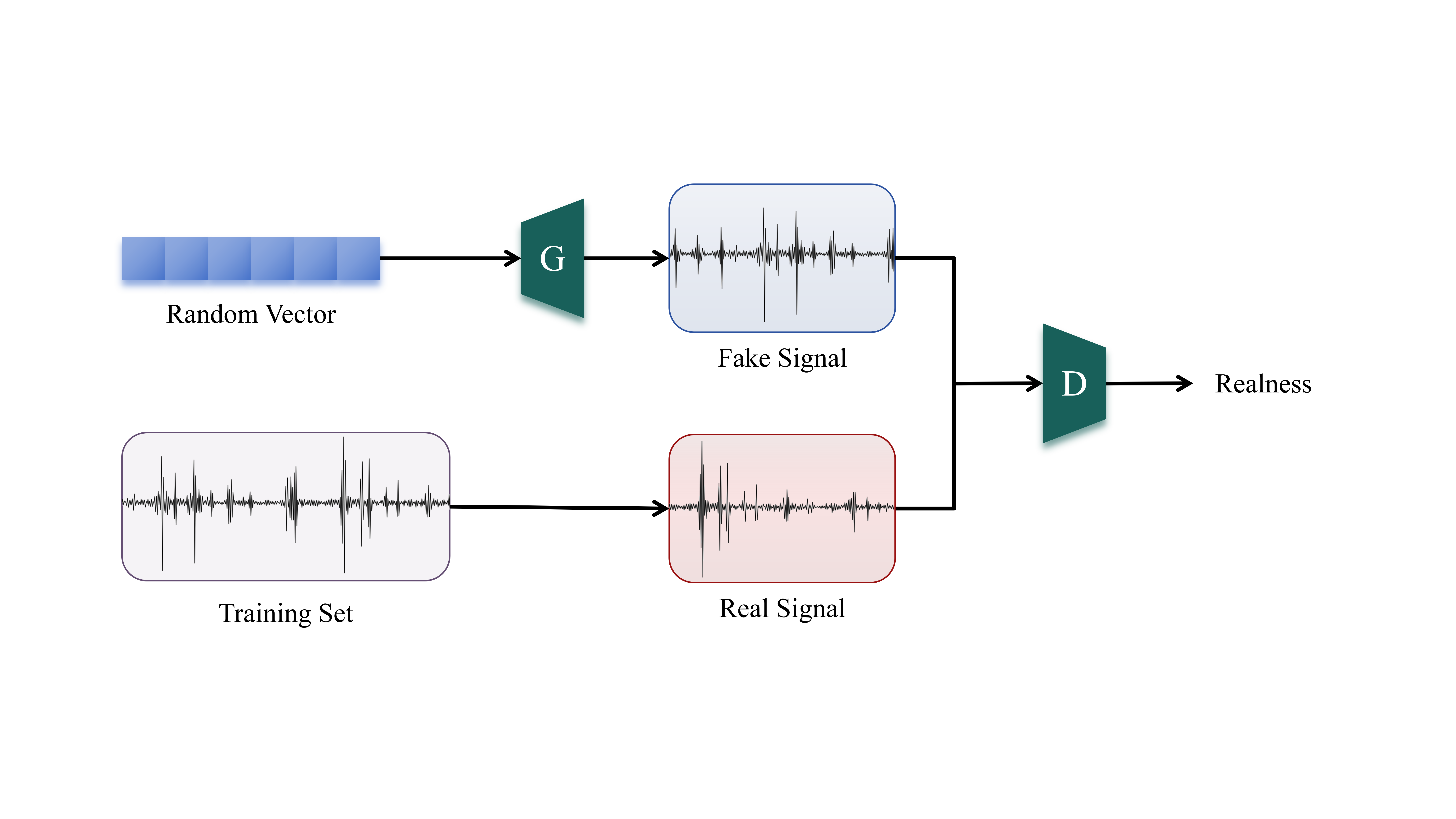}
  \caption{The flowchart of a conventional GAN.}
  \label{fig:conven_gan}
\end{figure}

Among all generative models, generative adversarial networks have been widely studied due to its outstanding ability to capture the data distribution and fast inference. A GAN usually consists of two components: a generator $G(\cdot)$ and a discriminator $D(\cdot)$. The generator aims to map a latent code $z$ into the data, while the discriminator tries to distinguish the generated image $G(z)$ from the real one $x$. The conventional GAN trains via a two-player game by optimizing the learning objective as follows:
\begin{align}
    \label{eq:equation1}
    \min_{G}\max_{D}V(D,G)=&\mathbb{E}_{x\sim p_{data}(x)}\left [ logD(x) \right ] + \\\notag &\mathbb{E}_{z\sim p_{z}(z)}\left [ log(1-D(G(z))) \right ]
\end{align}
where $p_{z}(z)$ and $p_{data}(x)$ represent a random latent representation and real data distribution respectively. The first term encourages the discriminator (D) to accurately classify real data points (high $logD(x)$ value). The second term encourages the generator (G) to produce realistic data that can fool the discriminator (low $log(1-D(G(z)))$ value). The flowchart of a conventional GANs can be seen in Fig.\ref{fig:conven_gan}.

Goodfellow et al.\cite{21} have proved that the optimal solution to this objective is that $G$ can eventually recover the distribution of source data and $D$ can not separate the generated images out from real images. However, the game between two players is not fair since in Eq. \ref{eq:equation1} that the gradient for optimizing all comes from $D$, which makes the discriminator a natural dominant. Thus the ideal solution is hard to reach in practice.

To address the limitations of conventional GANs, many GANs architectures have been proposed, among them, CLR-GAN\cite{CLR_GAN} is widely used due to its simple implementation and excellent performance. In this paradigm, the Generator and Discriminator are treated as inverse components. The latent code is used as a bridge to dynamically connect the Generator and the Discriminator, establishing a criterion for training the Discriminator. The Generator and Discriminator are then constrained by each other, making the game between the Generator and Discriminator more fair. However, the original CLR-GAN is unconditional, making it unsuitable for generating fault-type-specific samples required in bearing diagnosis.
To solve this problem, we design a conditional CLR-GAN (CCLR-GAN) that utilize cross attention to dynamically fuse the label information into generation process, this design makes the generative model can generate data points according to the demand labels stably, thus improving the quality of the generated samples. The implementation details can be seen in Sec. \ref{sec:clr_gan}.

\subsection{Deep Contrastive Learning}
\label{sec:contrastive}
Extracting recognizable feature representations from data is a fundamental task in deep learning. Many researchers have proposed that there is a strong positive correlation between the model's feature extraction ability and its performance. Among various training paradigms, contrastive learning has been proved to have a stronger ability to explore the relations inside the data without using supervised information (labels, regions,$\dots$). 

The workflow of a Neural Network can be divided into two parts: 1. a Feature Extraction Module and 2. a Task-Specified Module. The formulation of a Neural Network is:
\label{eq:contrastive}
\begin{align}
    &F(x) = \sigma(o(x)) \\
    &Z(x) = W(F(x))
\end{align}
where $o(\cdot)$ represents the filter function (convolution, attention), $\sigma(\cdot)$ is the activation function, $F(x)$ is the feature extractor, $W(\cdot)$ is the task-specified fully connected layer and $Z(x)$ is the linear classifier.

The formulation of Contrastive Learning can be represented as follows: We pair every data as $P:= ~ \langle x_{i},x_{j} \rangle$, and append a correlation label $y_{c}$ in the pair as:
\begin{align}
    &y_{c} = \left\{\begin{matrix}
  1,&y_{i} = y_{j} \\
  0,&y_{i} \ne y_{j}
\end{matrix}\right.
\end{align}
When the data pair is correlated, we denote it as $P^{+}$, when the data pair is uncorrelated, it is represented as $P^{-}$, the objective of contrastive learning can be expressed as:
\begin{align}
    &D(f(P^+)) \ll D(f(P^-))
\end{align}
where $f(\cdot)$ is the feature extraction module of a neural network, $D(\cdot)$ is a distance metric that evaluates the distance (similarity) between the feature spaces.

By the contrastive adaptation, we can generate a more compacted intra-class distribution and a more sparse inter-class distribution, thereby increasing the distance between the data, enhancing the extraction ability of the neural networks.
\begin{figure}
  \centering
  \includegraphics[width=\textwidth]{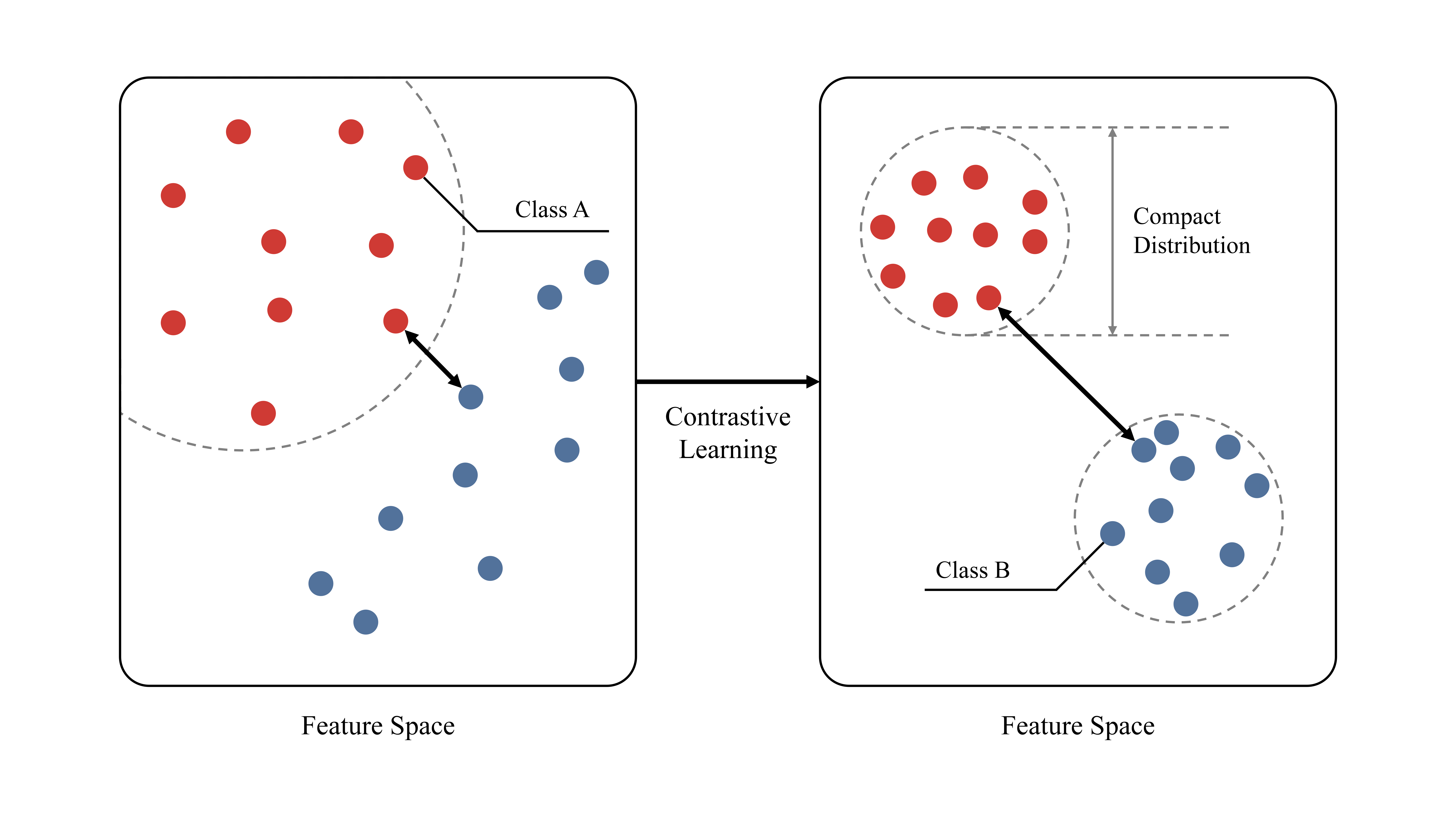}
  \caption{The feature space before (left) and after (right) the contrastive learning.}
  \label{fig:contrastive}
\end{figure}

As depicted in Fig.\ref{fig:contrastive}. The Contrastive Learning paradigm is suitable to limited data since it can explore the inner-correlation inside the training dataset. Thus the model can utilize the limited data more efficiently. The implementation detail of the contrastive learning based joint optimization mechanism are listed in Sec. \ref{sec:contras}.

\subsection{Fourier Transform in Machine Learning}
\label{sec:ft}
Fourier Transform is widely used in the spectral transform theory. It can transform signals from the time domain to the frequency domain, thereby decomposing the signal into components of different frequencies, which can better extract the frequency distribution of the signal and perform frequency domain analysis. The formulation of Fourier Transform is:
\begin{align}
    \mathcal{F}(\omega)=\int_{-\infty}^{+\infty} f(t)e^{-i\omega t}dt
\end{align}

After the frequency domain process/analysis, we can map the data back to time domain using:
\begin{align}
    f(t) = \frac{1}{2\pi}\int_{-\infty}^{+\infty}\mathcal{F}(\omega)e^{i\omega t}d\omega
\end{align}
where $f(t)$ is the time domain representation and $\mathcal{F}(\omega)$ is the frequency domain representation of the raw input.

It is proved in \cite{24} that a slight change in a single data point of a series of data will make the whole frequency domain different, thus the Fourier Transform can be seen as a global feature extractor. Consider a small perturbation to the frequency domain signal:
\begin{align}
    \mathcal{F}'(\omega) = \mathcal{F}(\omega) + \Delta F\cdot\delta(\omega-\omega_0)
\end{align}
where $\Delta F$ is perturbation amplitude and $\delta(\omega-\omega_0)$ is Dirac delta function which is non-zero at $\omega_0$.
By converting $\mathcal{F}'(\omega)$ back into time domain, we obtain:
\begin{align}
f'(t)&=\frac{1}{2\pi}\int_{-\infty}^{+\infty}[\mathcal{F}(\omega)+\Delta F\cdot\delta(\omega-\omega_0)]e^{i\omega t}\text{d}\omega\notag\\ 
&=\frac{1}{2\pi}\int_{-\infty}^{+\infty}\mathcal{F}(\omega)e^{i\omega t}\text{d}\omega+\frac{\Delta F}{2\pi}\int_{-\infty}^{+\infty}\delta(\omega-\omega_0)e^{i\omega t}\text{d}\omega\tag{Linearity of fourier transform}\\ 
&=f(t)+\frac{\Delta F}{2\pi}\cdot e^{i\omega_0t}\tag{Sifting property of the Dirac delta function}
\end{align}

This shows that a single modification in frequency domain leads to a global change in the time domain, thus if we can convert the bearing signal into frequency domain, the model can extract feature from a global perspective, enhancing model's ability to capture global patterns in the signal. 

Due to the lack of data, there is a strong need to make full use of the available data. While the previous feature extraction modules only have a local receptive field (convolution) or have a high computation demand (self-attention).

In this paper, we replace the conventional convolutional with fast fourier convolutional and design a 1D fourier convolution neural network (1D-FCNN) to improve the model's global feature extraction ability under limited data. It achieves better detection accuracy under similar parameter magnitude compared to conventional Convolutional Neural Networks. The implementation details are shown in Sec.\ref{sec:fft}.

\section{Proposed Method}
\label{sec:method}

In this section, we provide a comprehensive overview of the proposed DAC-FCF. The DAC-FCF is specifically designed to address the challenges of bearing fault diagnosis under limited data by integrating three key components: conditional data augmentation, contrastive feature learning, and global-aware feature extraction. 

Firstly, in Sec. \mbox{\ref{sec:overall}}, we present the overall architecture and data flow of the DAC-FCF, illustrating how its components interact to achieve high diagnostic accuracy. Next, in Sec. \mbox{\ref{sec:clr_gan}}, we introduce a novel conditional CLR-GAN (CCLR-GAN), which enhances training stability and enables controllable generation of fault-specific samples. This component addresses the issue of mode collapse and ensures the diversity of augmented data.

In Sec. \mbox{\ref{sec:contras}}, we detail the contrastive learning paradigm, including its loss function, which explores deeper relationships between limited training samples to improve feature discriminability. Finally, in Sec. \mbox{\ref{sec:fft}}, we demonstrate how FCNN are extended to 1-D FCNN for enhanced global awareness of input signals tailored for one-dimensional signals. This modification allows the model to capture both time-domain and frequency-domain features, improving its robustness and generalization ability.

\subsection{Overall Architecture}
\label{sec:overall}
\begin{figure}[h]
    \centering
    \includegraphics[width=\textwidth]{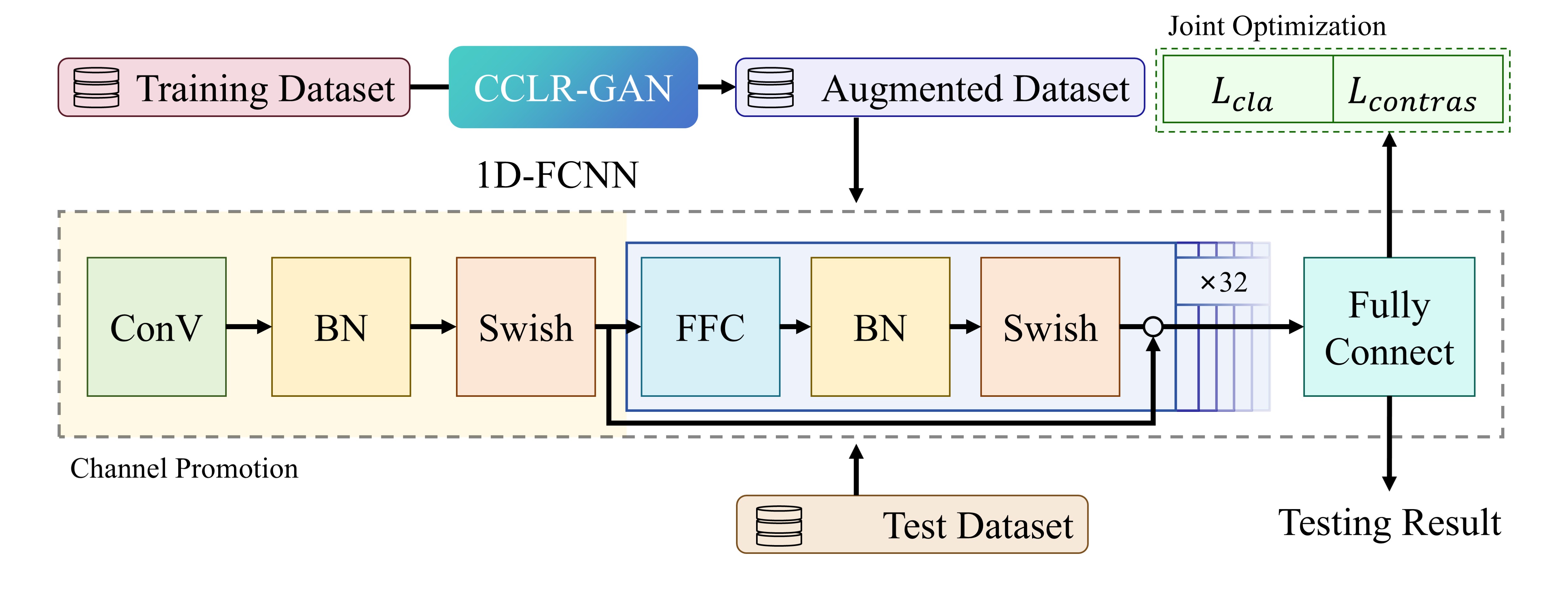}
    \caption{The overall architecture of DAC-FCF}
    \label{fig:dac_fcnn}
\end{figure}

The overall architecture of the proposed DAC-FCF can be seen in Fig.\ref{fig:dac_fcnn}. The overall architecture can be divided into the data generation stage and the fault diagnosis stage. Firstly, given the training dataset $D^{train}$, the CCLR-GAN is used to capture the distribution of data and generate more data for training. After the data augmentation stage, we can get an augmented dataset $D^{train}_{aug}$. Secondly, we split the training data $X$ into the local branch $X_l$ and global branch $X_g$ for fourier convolution that simultaneously extracts local and global information $X_l,X_g$ from the input signal. During feature extraction, we used ResNet-50 as the backbone for simplicity, and Swish is used as the activation function. Finally, we use a contrastive learning based joint optimization mechanism to reduce the feature space distance between positive samples while increasing the distance between positive and negative samples. 

\subsection{Conditional Consistent-Latent-Representation Generative Adversarial Networks}
\label{sec:clr_gan}

As we mentioned in Sec.\ref{sec:gan_intro}, in the conventional training paradigm GANs, the discriminator always trains better than the generator, making the gradient for the generator vanish. In practice, the discriminator can easily identify the fake samples from a relatively early stage of the training and maintain this advantage during the whole training process, making the generator hard to converge.

To address this problem, we utilize a recently proposed Consistent-Latent-Representation Generative Adversarial Networks (CLR-GAN)\cite{CLR_GAN} that utilize the latent code as an additional supervision signal for both Generator and Discriminator. Specifically, this paradigm treats the discriminator as an additional feature extractor $f(\cdot)$, by using the latent cosistency loss($\mathcal{L}_{CLR}$) and reconstruction loss($\mathcal{L}_{rec}$), the overall training objective of GANs can be formatted as:  
\begin{align}
    &\mathcal{L}_{D}' = \mathcal{L}_D + \lambda_{1}\mathcal{L}_{CLR} \\
    &\mathcal{L}_{G}' = \mathcal{L}_G + \lambda_{2}\mathcal{L}_{rec}
\end{align}
where $\mathcal{L}_{D}$ and $\mathcal{L}_G$ are the original loss function of the generator and discriminator in Eq.\ref{eq:equation1}. $\mathcal{L}_{rec}$ and $\mathcal{L}_{CLR}$ are the proposed real image reconstruction distance and consistent latent space distance. $\lambda_1$ and $\lambda_2$ are the two predefined weighting coefficient to control the strength of the proposed constrains. 

\begin{figure}[h]
    \centering
    \includegraphics[width=\textwidth]{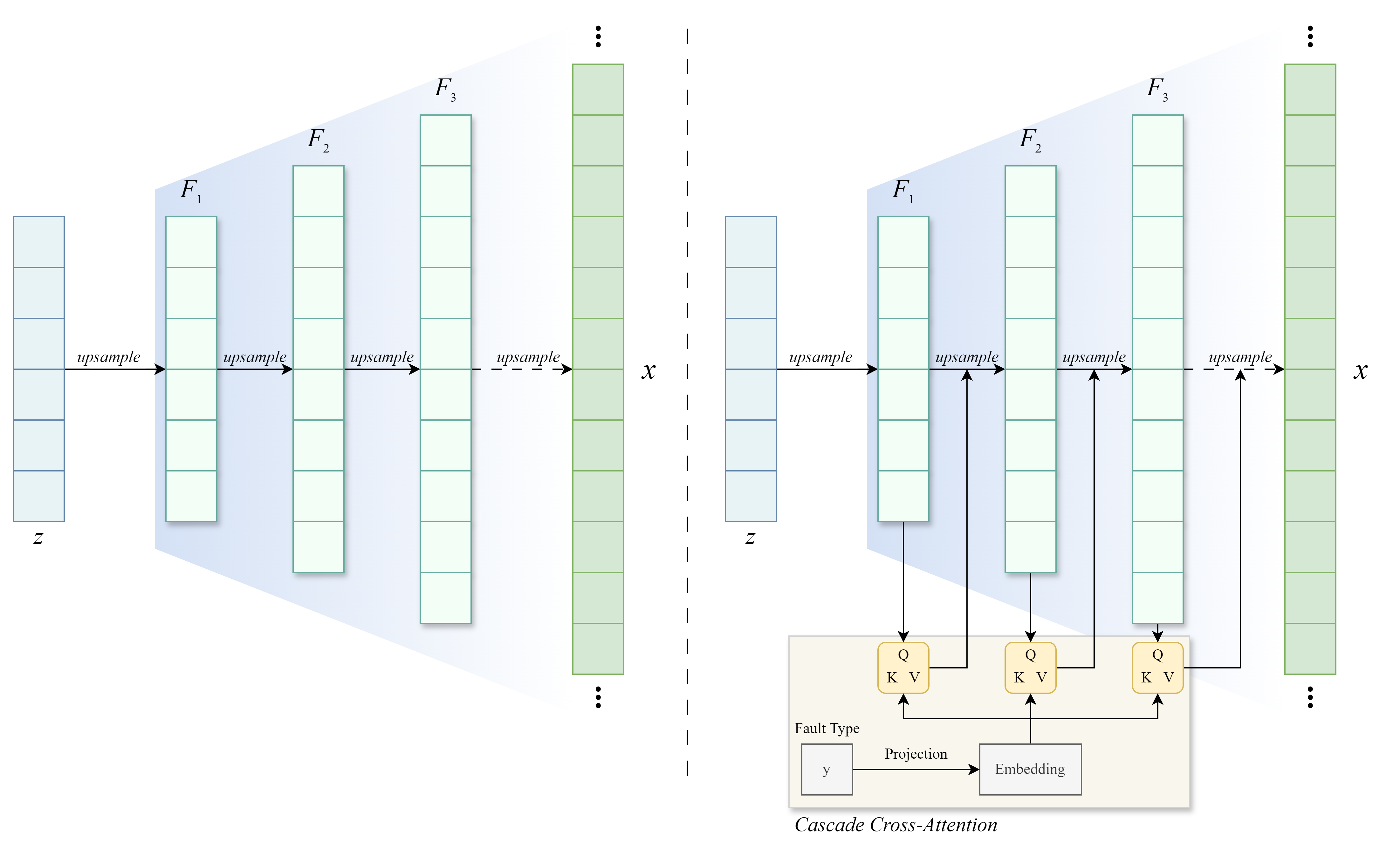}
    \caption{The overall architecture of CCLR-GAN}
    \label{fig:clr_gan}
\end{figure}

With the proposed objective, we can make the game between generator and discriminator more fair in the following two aspects: First, we can align the generator and the discriminator during training by $\mathcal{L}_{CLR}$, and the search space of generator is also constrained. Second, the reconstructed data serves as a bridge to connect the generator to the real distribution, thus making the generator make full use of the input data, leading to a more fidelity output. The detailed architecture of the proposed CLR-GAN is shown in Fig.~\ref{fig:clr_gan}

In the context of bearing fault diagnosis, the ability to generate high-quality, fault-specific samples is crucial for addressing the challenge of limited labeled data. Specifically, we aim to leverage Generative Adversarial Networks (GANs) to augment the training dataset by generating synthetic samples that correspond to specific fault types. However, the current structure of CLR-GAN, as an unconditional generative model, has a significant limitation: it can only generate samples according to the overall data distribution $P(x)$, without considering the conditional distribution $P(x\mid y)$, where $y$ represents the fault labels.

To enable CLR-GAN to generate fault-specific samples, we proposed a conditonal CLR-GAN (CCLR-GAN) that injects the label information into the generation process during training, thus making the generation process more controllable. Specifically, give the target label $y$ and random sampled latent code $z$, recall that the generation process can be considered as a cascade up sample using the generator, as illustrated in Fig.~\ref{fig:clr_gan} left. If we can inject the label information into the cascade generation process, we can eventually get a conditional CLR-GAN that generates the fault-specific samples.

Therefore, as depicted in Fig.~\ref{fig:clr_gan} right, we proposed a cascade cross-attention to inject the label information into the features, specifically, the target label $y$ first goes into a embedding layer to project it into a high dimensional semantic space $v$. This embedding captures the semantic representation of the fault type, which serves as a condition for guiding the generative process.

During the generation process, the cascade cross-attention operates at multiple scales of the generator's upsampling stages. At each upsampling step, the intermediate feature map $F_i$ produced by the generator is fused with the label embedding $v$ via a cross-attention module. The cross-attention mechanism computes the similarity between the feature map $F_i$ and the label embedding $v$, allowing the model to focus on regions of the feature map that are most relevant to the target fault type. This process can be formulated as:
\begin{equation}
    \label{cross-attention}
    F_{i}^{con} = \text{softmax}\left(\frac{F_iv^t}{\sqrt{d_v}}\right)v
\end{equation}
where $d_v$ is the dimensionality of the fault embedding. The output of the cross-attention $F_{i}^{con}$ ensures that the label information is dynamically integrated into the generation process.

This cascade cross-attention mechanism enables progressive guidance from coarse to fine-grained levels during the generation process. At earlier stages of upsampling, the cross-attention focuses on capturing global structural patterns related to the fault type. As the generation progresses to finer scales, the attention mechanism refines the details of the generated samples, ensuring that they align with the specific characteristics of the target fault. By iteratively applying cross-attention at each upsampling step, the generator can produce high-quality, fault-specific samples that exhibit both diversity and fidelity.

\subsection{Contrastive Learning based Joint Optimization}
\label{sec:contras}

As depicted in Sec.\ref{sec:problem_formulation}, it is not enough to just use class-wise classification methods since they often suffer from overfitting to specific pattern when the data are limited. So we propose to use pair-wise contrastive learning method for better feature representation. The core concept of contrastive learning is to squeeze the distribution space of positive samples and enlarge the distance between negative samples, thus making the whole space more separable.

To achieve the feature alignment using contrastive learning, we first pair the data $x \in D^{train}$ to create a new dataset $P^{train}=\{ (X_{i},Y_{i})|i=1,2,\dots,n_{pairs}\}$ where $X_{i}$ is the data pair $(x_{i},x_{j}), x\in D^{train}$ and $Y_{i}$ is defined as:
\begin{equation}
    Y = \left\{\begin{matrix}
  1,& y_{i}=y_{j}\\
  0,& y_{i}\ne y_{j}
\end{matrix}\right.
\end{equation}

During every mini-batch, the data pair is fed into the feature extractor of the network to get the features:
\begin{align}
    D(x_{i}) = F(x_{i}) \notag\\
    D(x_{j}) = F(x_{j})
\end{align}

We can then calculate the similarities between the data pair using various distance metrics. For simplicity, we used cosine similarity to evaluate the the distance between the data pair:
\begin{equation}
    S_{ij} = \frac{D(x_{i})\cdot D(x_{j})^T}{||D(x_{i})||_2 \cdot||D(x_{j})||_2 }
\end{equation}

Finally, we can use a regular Cross-Entropy Loss to calculate the contrastive loss:
\begin{equation}
    \mathcal{L}_{con}(S_{ij},Y_{i}) = -(Y_{i}\cdot \log_{}{S_{ij}} + (1-Y_{i})\cdot \log_{}{(1-S_{ij})} )
\end{equation}

By this contrastive loss, we can gradually enlarge the inter-class distance and shrink the intra-class distance, thus making the feature space more compact. The classifier then can separate the data points easier.

\subsection{Fast Fourier Transform Convolutional Neural Networks for Global Aware Feature Extraction}
\label{sec:fft}

As mentioned in \ref{sec:ft}, conventional Convolutional Neural Networks (CNNs) have a limited receptive field. The convolutional kernel can only recept the data points inside the detection range, preventing the model to have global awareness. Furthermore, when data size is the bottleneck, it is significant to make full use of the available data, thus using the convolutional neural networks alone is not enough. 

To solve this problem, inspired by \cite{20}, we utilize the Fourier Convolutional Neural Networks and convert it to a 1-D global aware Fourier Convolutional Neural Networks (1D-FCNN) tailored specifically for one-dimensional vibration signals.Our proposed 1D-FCNN incorporates two key improvements over existing Fourier Convolutional Networks: (1) dimensionality reduction using 1×1 convolutions to reduce computational complexity, and (2) adaptive convolutional strides that first employ large strides to capture coarse-grained global features, followed by small strides to refine fine-grained details.

Specifically, the 1D-FCNN is a dual-path feature fusion network. The convolutional layer is divided into two path: a local aware path that uses the conventional convolution and a global aware path that uses fourier transform (FT) to convert the time-domain signals to frequency-domain to get the global frequency-domain features. Information exchange is utilized in each layer.

Consider an input $X\in R^{N\times L}$ where $N$ represents number of samples and $L$ is the length of input signal. We first expand the channel from $1$ to $C$ using a regular $1\times 1$ convolutional layer. Then we split $X$ into $X=\{ X_l,X_g \}\in R^{N\times \frac{C}{2} \times L}$. $X_l,X_g$ are put into the local branch and global branch, respectively. 

\begin{figure}
  \centering
  \includegraphics[width=\textwidth]{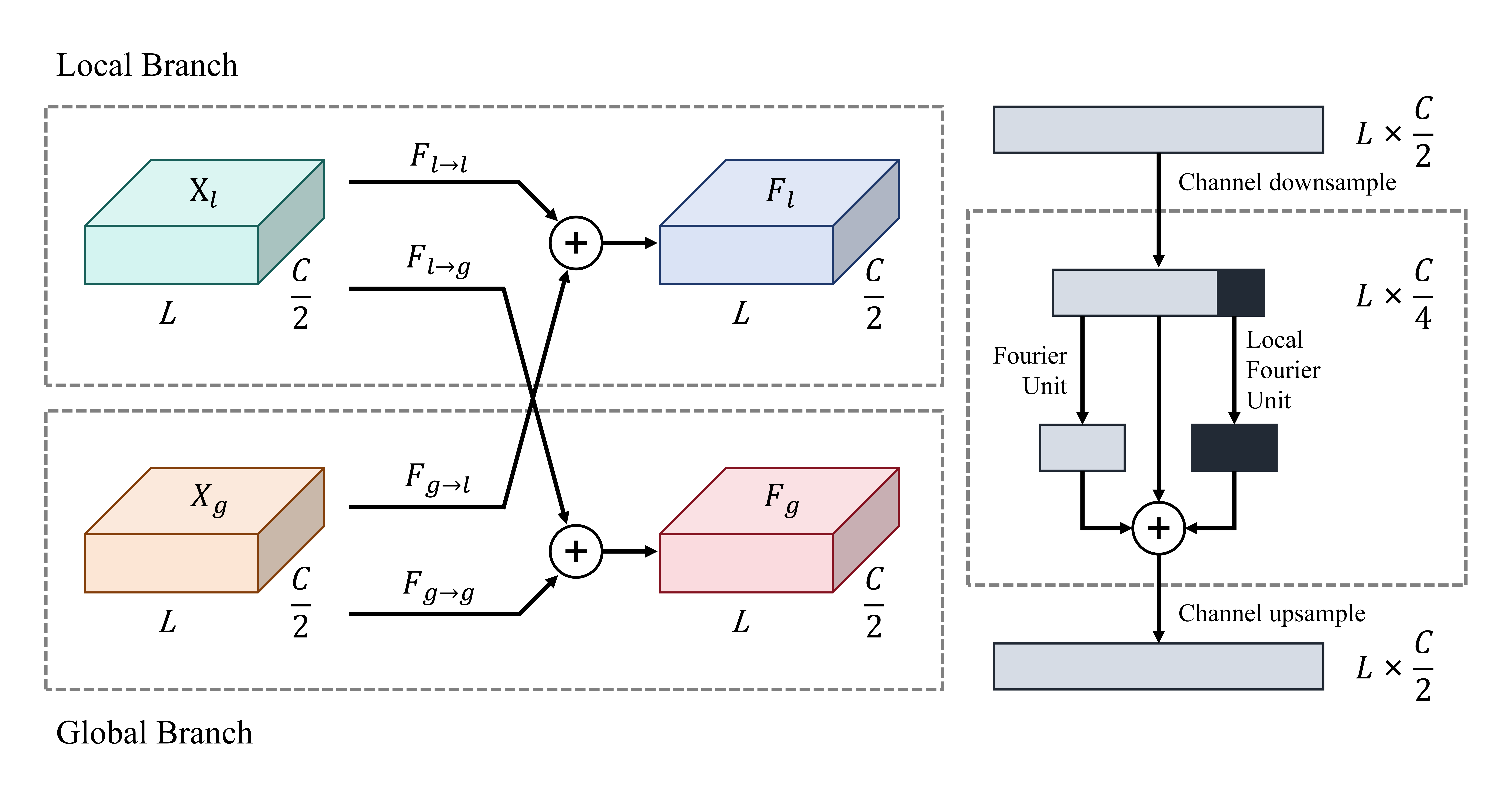}
  \caption{The overall structure of Fourier Convolutional Neural Network.}\label{fig:ffc}
\end{figure}

Denote that the extracted features of both branches after the Fourier convolutional neural networks as $F_g,F_l$, these two features can be expressed as:
\begin{align}
    &F_g(X_g) = F_{g\to g}(X_g) + F_{l\to g}(X_l) \\
    &F_l(X_l) = F_{l\to l}(X_l) + F_{g\to l}(X_g)
\end{align}
where $F_{g\to g}$ aims to extract the global aware features using Fourier Convolution and $F_{l\to l}$ captures local aware features using conventional CNNs, $F_{l\to g}$ and $F_{g\to l}$ obtain path feature fusion using a regular CNNs. The main component in the proposed method is the Fourier Convolutional Networks, the implementation detail is depicted in Fig.\ref{fig:ffc}.

To further optimize the model's performance for one-dimensional signals, we introduce adaptive convolutional strides in the global-aware path. Specifically, the convolutional kernel initially employs a large stride to capture coarse-grained global features across the entire signal. Subsequently, a small stride is used to refine fine-grained details, enabling the model to better exploit both long-range dependencies and localized patterns. This strategy ensures that the model effectively captures hierarchical information from the input signal, enhancing its global awareness.

To reduce computation, the input $X_{g}$ first gets through a $1\times 1$ convolutional block to cut the channels in half. Then the data are fed into the Fourier Unit for global feature extraction and a Local Fourier Unit to extract semi-global frequency-domain features. Finally, the input and the output of Fourier Unit and Local Fourier Unit are added together and go through a $1\times 1$ convolutional layer to double the channel as the output.

By utilizing the 1D-FCNN, we successfully extract features from both time-domain and frequency-domain to get more accurate representations. It also enables the model to get global-awareness on the data, thus using the available data more effectively under limited data scenarios. Experiment results illustrate the effectiveness of utilizing Fourier Convolutional Neural Networks.

\section{Experiments and Discussion}
\label{sec:exp}

In this section, we perform abundant experiments to prove the effectiveness of the proposed method. We first give a detailed information about the two training datasets. Then we introduce the training hyperparameters (optimizer, learning rate) of the proposed DAC-FCF in Sec.\ref{sec:setups}. In Sec.\ref{sec:accuracy}, we validation the effectiveness of the proposed DAC-FCF compared to other methods. In Sec.\ref{sec:ablation}, we conduct some ablation experiments to show the necessity of the proposed three techniques. At last, in Sec.\ref{sec:fairer}, we proved the proposed CCLR-GAN can lead to a fairer game and generates data more precisely.

\subsection{Datasets}
\label{sec:dataset}

\begin{figure}[!h]
  \centering
  \includegraphics[width=\textwidth]{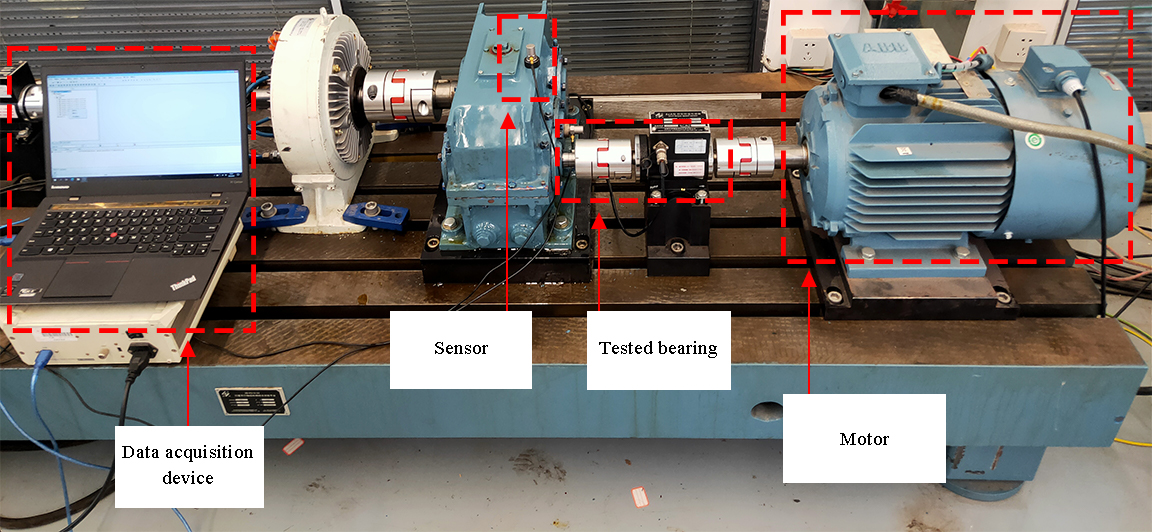}
  \caption{Experimental environment of the self-collected test bench.}
  \label{fig:hebut}
\end{figure}

In order to test the effectiveness of the proposed DAC-FCF, we use the Case Western Reserve University (CWRU) bearing fault dataset\cite{28} and a self collected two-stage drive bearing fault diagnosis test bench as our dataset. The testing environment of the self collected bench is shown in Fig.\ref{fig:hebut}. The detailed information of the dataset is as follows.

\subsubsection{CWRU Dataset}
The CWRU dataset includes vibration signals of three types of faults: inner race, rolling ball, and outer race. The dataset uses data from 12 kHz drive end rolling bearings, collected with damage diameters of 0.1178 mm and 0.3556 mm. Seven types of vibration signals including normal condition is collected. The detailed structure of CWRU dataset is listed in Tab.\ref{tab:cwru_dis}.

\begin{table}[]
\caption{Description of labels and fault types of CWRU dataset}
    \centering
    \begin{tabular}{|c|c|c|}
    \hline
        \textbf{Health conditions} &\textbf{Label}&\textbf{Fault diameters/inch}  \\
        \hline
         Normal&0&- \\
        \hline
        Inner Raceway&1&0.007 \\
        \hline
        Inner Raceway&2&0.014 \\
        \hline
        Inner Raceway&3&0.021 \\
        \hline
        Outer Raceway&4&0.007 \\
        \hline
        Outer Raceway&5&0.014 \\
        \hline
        Outer Raceway&6&0.021 \\
        \hline
        Ball&7&0.007 \\
        \hline
        Ball&8&0.014 \\
        \hline
        Ball&9&0.021 \\
        \hline
    \end{tabular}
    
    \label{tab:cwru_dis}
\end{table}
\subsubsection{Self-Collected Test Bench}

Since the CWRU dataset is already well studied and it only contains few working conditions in it. To better validate the proposed DAC-FCF, we turn to collect a more challenging dataset which uses the middle shaft bearing of the two-stage gearbox, and the sample frequency is 96kHz at the drive end. The collected data labels include six different health conditions: health, inner fault, outer fault, roller fault, inner roller fault and outer roller fault.

We combine three rotation speed and three working load to get 9 different working conditions to enhance the variety of data.  Before the experiment begins, the test bench was warmed up for 30 minutes, and then operated for 5 minutes after switching between different operating conditions to obtain stable signals. The correspondence between label and fault type is listed in Tab.\ref{tab:self_dis}

\begin{table}[]
\caption{Description of labels and fault types of Self-Collected dataset}
    \centering
    \begin{tabular}{|c|c|c|c|}
    \hline
        \textbf{Health conditions} &\textbf{Label}&\textbf{Rotation speed/rpm}&\textbf{Load/A}  \\
        \hline
         Normal&0&1200/1500/1800&0/0.5/1 \\
        \hline
        Inner Raceway&1&1200/1500/1800&0/0.5/1 \\
        \hline
        Outer Raceway&2&1200/1500/1800&0/0.5/1 \\
        \hline
        Roller Fault&3&1200/1500/1800&0/0.5/1 \\
        \hline
        Inner Roller&4&1200/1500/1800&0/0.5/1 \\
        \hline
        Outer Roller&5&1200/1500/1800&0/0.5/1 \\
        \hline
    \end{tabular}
    
    \label{tab:self_dis}
\end{table}

For both datasets, we used the sliding window method to extract 1024 non overlapping data points as one piece of data. The data size of both datasets is 20, 50, 100, 150, 200 respectively to test the performance of different models under various conditions.

\subsection{Experiment Setups}
\label{sec:setups}
As illustrated in Fig.\ref{fig:dac_fcnn}. In the data generation stage, the optimizer of both generator and discriminator is Adam. The learning rate for both generator and discriminator is 0.0001. During the fault diagnosis stage, the optimizer is Adam as well, and the learning rate during the fault diagnosis stage is 0.0002. After the CCLR-GAN is trained, we generate 500 samples each class. The batch size for data generation is 32 and the batch size for fault diagnosis is 64. For each dataset, we validate the diagnose accuracy under 5 different sample sizes, where sample size are defined as the number of training samples available each class during training. The full list of hyper parameters are listed in Tab.\ref{tab:hyperparam}

\begin{table}[]
\caption{Training Parameters of DAC-FCF}
    \centering
    \begin{tabular}{|c|c|c|}
    \hline
        Hyperparameters & CCLR-GAN & Classification  \\
        \hline
         Optimizer&Adam&Adam \\
        \hline
        Data length&1024&1024 \\
        \hline
        Noise dim&100&\diagbox{}{} \\
        \hline
        Batch size&32&64 \\
        \hline
        lr&1e-4&2e-4 \\
        \hline
    \end{tabular}
    
    \label{tab:hyperparam}
\end{table}

\subsection{Diagnosis accuracy}
\label{sec:accuracy}

We first test the diagnosis accuracy of CWRU dataset and self-collected testing bench under different sample size. We compare our method with mostly used TCA\cite{17}, CNN\cite{CNN}, DACNN\cite{DACNN}, DDC\cite{DDC} and RNN based FDGRU\cite{FDGRU}. Transformers are proved to have strong performance when dealing with long sequences, but the capability of it under limited data are under explored recently, so we also compare with the recently proposed transformer based SiT\cite{SIT}, the results are shown in Table.\ref{tab:cwru_acc}. and Table.\ref{tab:hgd_acc}. The proposed DAC-FCF outperforms other baselines when the sample size is extremely small (20), which indicates that the proposed DAC-FCF can benefit from the generated samples and can discover more meaningful patterns between different kinds of faults. Our proposed model also achieved the best result on the average accuracy. Note that the diagnosis accuracy of DDC is high since it uses data from different domains as extra information to train the model, which is usually unavailable in real world. Despite DDC with additional data, our model consistently achieves the second best results compared to other recently proposed models. And when the extra data is removed from the training process, we can see a significantly drop on the DDC model (DDC w/o extra), the DDC model can only achieve 53\% accuracy without extra information while our DAC-FCF can still achieve more than 90\%, further proving the effectiveness of the proposed method.

Another thing to mention is that CWRU is usually considered as a relatively easy dataset. However, in our settings, when there only consists tens of samples per class, CWRU dataset becomes more challenging. As is shown in Table.\ref{tab:cwru_acc}, in the most extreme case (20), all of the baselines fail to learn recognizable features. While our proposed DAC-FCF remains a competitive result. This indicates that in our settings, CWRU dataset becomes more difficult and can serve as a reliable dataset.

On the more challenging self-collected test bench, the proposed DAC-FCF can achieve an average accuracy of 80.34\%, while other methods without using extra information can only reach less than 50\%. This phenomena further proves the importance of data, when enough data is provided, the model can generalize way better than the augmented dataset. Another thing worth noting is transformer architecture are more vulnerable to data scarcity, when sample size is small, the SiT suffers from severe overfitting, leading to a random guess during validation.

\begin{figure}[t]
    \centering
    \includegraphics[width=\textwidth]{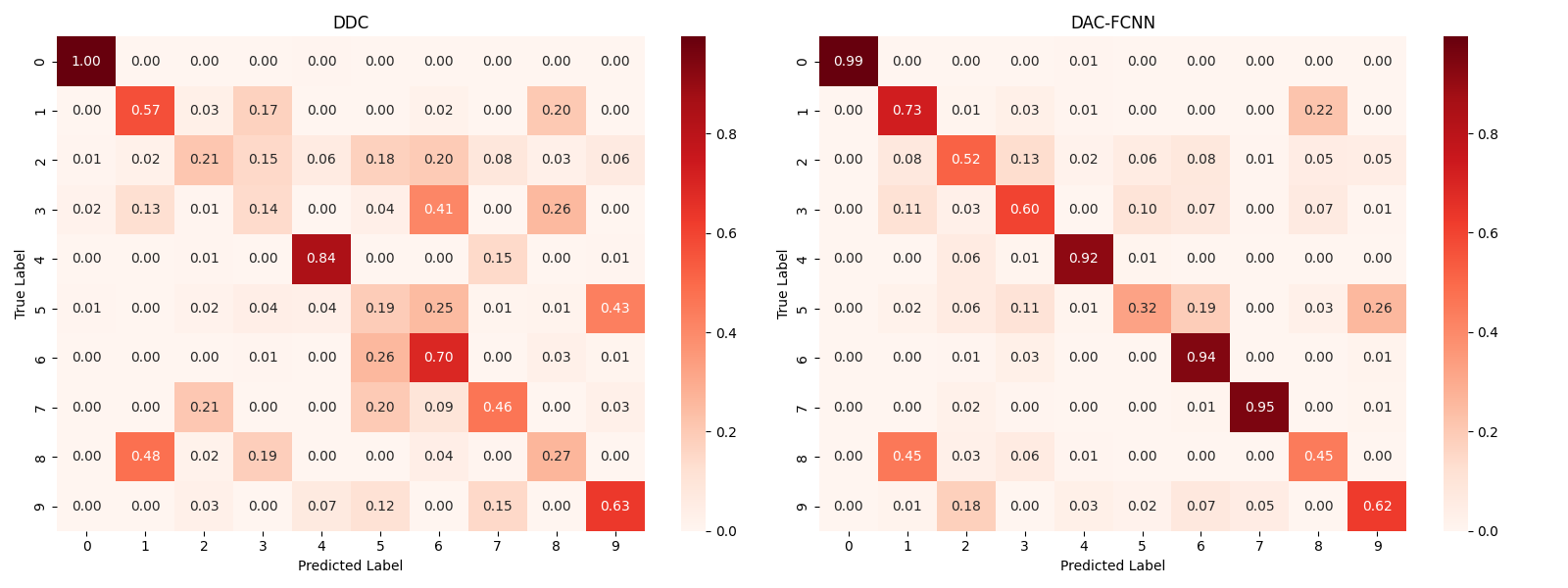}
    \caption{Per-class diagnosis accuracy of DDC (left) and the proposed DAC-FCF (right).}
    \label{fig:confusion}
\end{figure}

To better analysis the per-class diagnosis accuracy of the proposed model, we plot the confusion matrix on CWRU dataset of the previous state-of-the-art DDC model and the proposed DAC-FCF. From Fig.\ref{fig:confusion}, due to the long-tail property of bearing fault data, the DDC model performs well only on the head class (normal bearings), while our proposed model better captures the inner pattern of the bearing data and gets reasonably performance among all the classes. 
\begin{table*}[]
\caption{Diagnosis accuracy of CWRU dataset under different sample size.}
    \centering
    \begin{tabular}{|c|c|c|c|c|c|c|}
    \hline
        \diagbox{Method}{Sample Size} & 20&50&100&150&200&Average \\
        \hline
         TCA&19.6\%&19.55\%&18.85\%&18.45\%&18.45\%&18.98\% \\
         \hline
         CNN&34.44\%&41.63\%&59.92\%&63.76\%&69.14\%&53.78\% \\
         \hline
         DACNN&31.68\%&\underline{50.65}\%&65.97\%&71.05\%&71.78\%&58.22\%\\
         \hline
         DDC&\underline{41.64}\%&50.52\%&\textbf{92.67\%}&\textbf{98.58\%}&\textbf{99.76\%}&76.64\%\\
         \hline
         DDC w/o extra&11.15\%&16.59\%&26.24\%&55.10\%&53.21\%&32.45\%\\
         \hline
         SiT&14.59\%&15.77\%&43.38\%&50.32\%&59.79\%&36.77\%\\
         \hline
         FDGRU&21.49\%&28.65\%&41.06\%&46.32\%&50.17\%&37.54\%\\
         \hline
         DAC-FCF (\textbf{Ours})&\textbf{74.02\%}&\textbf{84.97\%}&\underline{87.73}\%&\underline{87.96}\%&\underline{90.04}\%&\textbf{84.94\%}\\
         \hline
    \end{tabular}
    \label{tab:cwru_acc}
\end{table*}

\begin{table*}[]
\caption{Diagnosis accuracy of self-collected dataset under different sample size.}
    \centering
    \begin{tabular}{|c|c|c|c|c|c|c|}
    \hline
        \diagbox{Method}{Sample Size} & 20&50&100&150&200&Average \\
        \hline
         TCA&19.3\%&20.55\%&18.85\%&19.45\%&22.45\%&20.12\% \\
         \hline
         CNN&26.02\%&36.35\%&39.20\%&39.40\%&46.02\%&37.39\% \\
         \hline
         DACNN&24.79\%&35.84\%&38.47\%&45.30\%&42.27\%&37.33\%\\
         \hline
         DDC&53.80\%&\textbf{85.76\%}&\textbf{93.19\%}&\textbf{94.46\%}&\textbf{96.39\%}&\textbf{84.72\%}\\
         \hline
         DDC w/o extra&41.31\%&49.70\%&46.83\%&48.09\%&54.65\%&48.12\%\\
         \hline
         SiT&21.93\%&17.94\%&19.37\%&29.58\%&34.40\%&24.64\%\\
         \hline
         FDGRU&21.03\%&23.33\%&30.30\%&34.82\%&39.80\%&37.36\%\\
         \hline
         DAC-FCF (\textbf{Ours})&\textbf{63.24\%}&\underline{66.23}\%&\underline{83.06}\%&\underline{84.98}\%&\underline{90.08}\%&\underline{80.34}\%\\
         \hline
    \end{tabular}
    
    \label{tab:hgd_acc}
\end{table*}

\subsection{Ablation study}
\label{sec:ablation}
\begin{figure}[t]
    \centering
    \includegraphics[width=\textwidth]{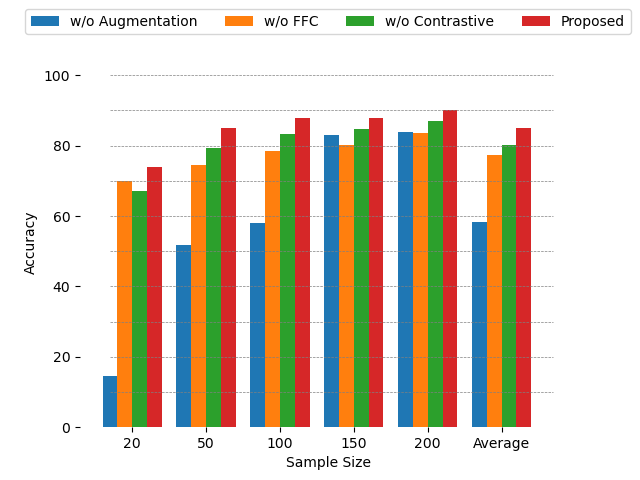}
    \caption{Accuracy with/without CCLR-GAN, contrastive learning, and Fourier Convolution under different sample sizes.}
    \label{fig:ablation}
\end{figure}

In order to validate the effectiveness of the proposed components, we conduct extensive ablation experiments in this section. The experiment results are shown in Fig.\ref{fig:ablation}. From the experiment results, it is obvious that the data augmentation method plays an important role in the performance of the model. When we disable the augmentation method, the diagnosis accuracy drastically dropped 50\% at most, which implies that the data quality is the most important factor when the sample size is small, with the sample size growing, the effectiveness of augmentation methods gradually becomes weaker, since the sample size is enough to guide the model to extract representative features.

The Fourier Convolution and Contrastive learning paradigm are also important to achieve a hight accuracy, when utilizing contrastive learning, the model can better delve the intra-relationship between data pairs, achieving a clearer boundary between different classes, while the Fourier Convolution helps to obtain long range awareness of the data, enabling the model to make judgments based on overall data features, improving the robustness of the proposed model.

\subsection{Towards a fairer game}
\label{sec:fairer}

\begin{figure}[htbp]
\centering
\subfigure{
\includegraphics[width=6.5cm]{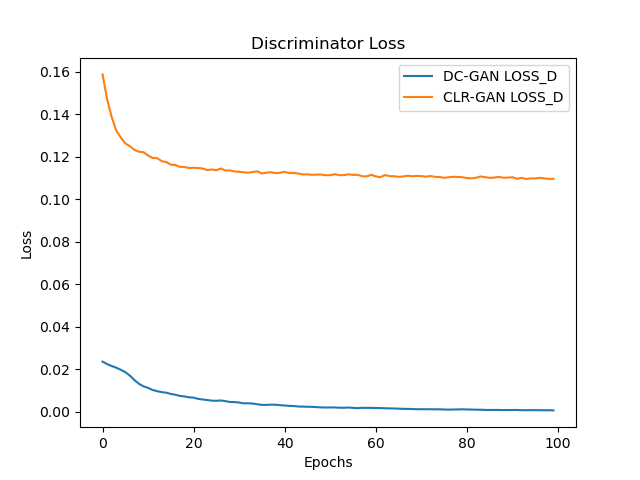}
}
\hfill
\subfigure{
\includegraphics[width=6.5cm]{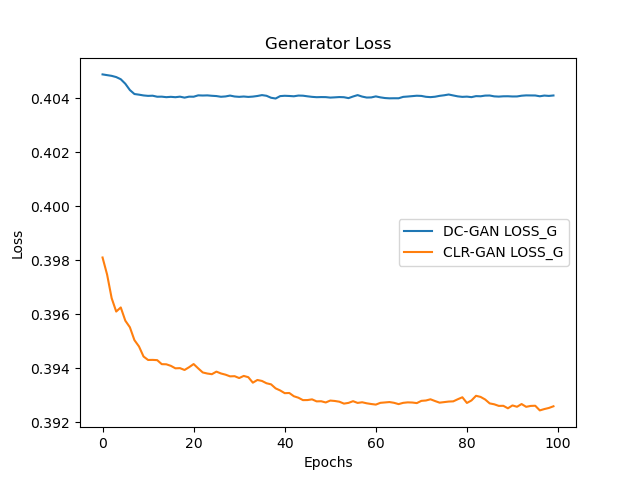}
}

\caption{The training loss of the proposed CCLR-GAN and DC-GAN}\label{fig:fig_fair}
\end{figure}

In this section, we demonstrate that our proposed CCLR-GAN can make the game between the generator and discriminator more fair compared to the conventional training paradigm. Specifically, we collected the training loss of generator and discriminator using the same experimental settings. The visualization result is shown in Fig.\ref{fig:fig_fair}. It is obvious that when using the conventional training paradigm, the loss of the discriminator quickly dropped to below 0.02, causing the gradient vanishing problem, making the generator unable to convergence. While the proposed CCLR-GAN can make the game between generator and discriminator more fair, the discriminator can offer valuable gradient during training, thus the generate ability of the generator can be further improved.

\begin{table}
\caption{Diagnosis Accuracy of CWRU dataset under different types of GAN architectures.}
\centering
\begin{tabularx}{\textwidth}{|l|*{6}{>{\centering\arraybackslash}X|}}
\hline
\diagbox{Method}{Sample Size} & 20 & 50 & 100 & 150 & 200 & Average \\
\hline
DCGAN+FCNN & 17.13\% & 61.32\% & 71.42\% & 79.34\% & 87.94\% & 63.43\% \\
\hline
CCLR-GAN+FCNN & 74.02\% (+56.89\%) & 84.97\% (+23.65\%) & 87.73\% (+16.31\%) & 87.96\% (+8.62\%) & 90.04\% (+2.1\%) & 84.94\% (+21.51\%) \\
\hline
\end{tabularx}
\label{tab:gan_compare}
\end{table}

Next, we carry out experiments to prove the stability of the proposed CCLR-GAN under limited data, we use DCGAN and CCLR-GAN as two augmentation methods to generate new samples, the experimental results are depicted in Table.\ref{tab:gan_compare}. It is obvious that when the sample size is large (200), both DCGAN and CCLR-GAN perform well on generating new samples, however, as the sample size shrinks, the performance gap between DCGAN and the proposed CCLR-GAN becomes significant. This phenomenon implies that when the sample size grows smaller, the conventional GAN architecture suffers from mode collapse, and it can not generate meaningful samples, thus harming the fault diagnosis network. Conversely, the proposed CCLR-GAN can precisely capture the distribution of data regardless of sample size. This further proves the effectiveness and stability of our proposed CCLR-GAN.

\section{Conclusion and Future Work}
\label{sec:conclusion}

In this paper, we propose DAC-FCF, a novel bearing fault diagnosis framework tailored for limited-data scenarios, with three core contributions:
Firstly, a CCLR-GAN architecture is introduced to address the instability in traditional GAN training. By dynamically balancing generator-discriminator optimization, CCLR-GAN achieves superior data augmentation quality compared to conventional methods.
Secondly, a contrastive learning paradigm is designed to exploit inter-sample relationships. By constructing positive/negative pairs and maximizing feature consistency within pairs, our approach enhances discriminative feature extraction under data scarcity.
Lastly, an enhanced fourier convolutional neural networks is developed to capture multi-scale global patterns. This module outperforms standard CNNs in extracting long-range dependencies from bearing fault vibration signals. Extensive experiments proved the effectiveness of the proposed DAC-FCF under limited data.

While GANs demonstrates superior performance in limited-data scenarios, its scalability to large scale datasets remains constrained due to the inherent instability of GAN training and the computational burden of adversarial learning. A future work of DAC-FCF is to utilize different network architectures to further enhance the model's receptive field and to extract features of different levels. Although Fourier convolution can to some extent extract global information, there are also some methods using cluster based or graph based methods to enhance the interaction of samples\cite{wh2,wh3,wh4}. Thus making the model able to extract more general features, thereby improving the model's performance across different domains. Another important aspect is to incorporate privacy-preserving techniques into the design and application of the model\cite{wh5,wh6,wh7,wh8,wh9,wh10}. By employing privacy-preserving techniques, we can effectively extract useful information while ensuring data security, thereby enhancing the robustness and generalization capabilities of the model.

\bibliographystyle{plainnat}
\bibliography{bibtex}

\end{document}